\pdfoutput=1

\documentclass[11pt]{article}

\usepackage{EMNLP2022}

\usepackage{times}
\usepackage{latexsym}

\usepackage[T1]{fontenc}

\usepackage[utf8]{inputenc}

\usepackage{microtype}

\usepackage{graphicx}
\usepackage{subfigure}
\usepackage{multirow}

\usepackage{amsmath,amssymb,amsfonts}
\usepackage{mathtools}

\usepackage{booktabs}
\usepackage{tabularx}

\usepackage{algorithmic}
\usepackage{graphicx}
\usepackage{textcomp}
\usepackage[ruled,vlined]{algorithm2e}
\usepackage{arydshln}
\usepackage{blindtext}
\usepackage{hyperref}
\usepackage{booktabs}
\usepackage{multirow}

\makeatletter
\newcommand{\removelatexerror}{\let\@latex@error\@gobble}
\makeatother

\usepackage{enumitem}
\setenumerate[1]{itemsep=2pt,partopsep=0pt,parsep=\parskip,topsep=5pt}
\setitemize[1]{itemsep=2pt,partopsep=0pt,parsep=\parskip,topsep=5pt}
\setdescription{itemsep=2pt,partopsep=0pt,parsep=\parskip,topsep=5pt}

\usepackage{ntheorem}
\theoremseparator{:}

\usepackage{inconsolata}

\title{A Generative Model for Joint Multiple Intent Detection and Slot Filling }

\author{
    Liz Li\textsuperscript{1,}, Wei Zhu\textsuperscript{2,}\thanks{\ \ Corresponding author. For any inquiries, please contact: michaelwzhu91@gmail.com. }\\
    \small \textsuperscript{1}\ DataSelect AI, Xuhui, Shanghai, China \\
    \small \textsuperscript{2}University of Hong Kong, Hong Kong, HK, China
}

\begin{document}
\maketitle
\begin{abstract}
In task-oriented dialogue systems, spoken language understanding (SLU) is a critical component, which consists of two sub-tasks, intent detection and slot filling.
Most existing methods focus on the single-intent SLU, where each utterance only has one intent.
However, in real-world scenarios users usually express multiple intents in an utterance, which poses a challenge for existing dialogue systems and datasets.
In this paper, we propose a generative framework to simultaneously address multiple intent detection and slot filling.
In particular, an attention-over-attention decoder is proposed to handle the variable number of intents and the interference between the two sub-tasks by incorporating an inductive bias into the process of multi-task learning.
Besides, we construct two new multi-intent SLU datasets based on single-intent utterances by taking advantage of the next sentence prediction (NSP) head of the BERT model.
Experimental results demonstrate that our proposed attention-over-attention generative model achieves state-of-the-art performance on two public datasets, MixATIS and MixSNIPS, and our constructed datasets.

\end{abstract}

\section{Introduction}
\label{sec:introduction}

In modern task-oriented dialogue systems, spoken language understanding (SLU) is a neccessary component to capture the semantics of user queries. Typically, SLU is comprised of two sub-tasks, namely intent detection and slot filling\cite{tur2011spoken}. Take the utterance "\emph{Put Ramy Ayach on latin pop rising}" as an example, the intent detection task is to recognize the intent \texttt{AddToPlaylist}, while the slot filling task is to find the values for slots, \emph{i.e.} \emph{Ramy Ayach} as the \texttt{artist} and \emph{latin pop rising} as the \texttt{playlist}. In prior work\cite{DBLP:conf/asru/XuS13, DBLP:conf/interspeech/RavuriS15, DBLP:conf/interspeech/RaymondR07,Wu2018UnsupervisedFL,Oord2018RepresentationLW,Tian2020ContrastiveMC,He2020MomentumCF,Chen2021ExploringSS,Cui2023UltraFeedbackBL,wang2024ts,yue2023-TCMEB,yue2024tcmbench,Zhang2023LearnedAA,2023arXiv230318223Z,Xu2023ParameterEfficientFM,Ding2022DeltaTA,Xin2024ParameterEfficientFF,qin2023chatgpt,PromptCBLUE,text2dt_shared_task,zhu2024text2mdt,zhu_etal_2021_paht,Li2023UnifiedDR,Zhu2023BADGESU,Zhang2023LECOIE,Zhu2023OverviewOT,guo-etal-2021-global,zhu-etal-2021-discovering,Zheng2023CandidateSF,Sun2020MedicalKG,Zhang2023NAGNERAU,Zhang2023FastNERSU,Wang2023MultitaskEL,Zhu2019TheDS,zhu2021leebert,Zhang2021AutomaticSN,Wang2020MiningIH,li2025ft,leong2025amas,zhang2025time,yin2024machine,zhu2026mrag,zhu2026evaluatechatgpt,liu2024alora,zhu-2021-autorc,autotrans,zhang2024milora,li2019pingan,zhu2021autonlu,zhu2021lex,zheng2024sca,zhang2025adapting,zheng2024chimera,zhu2021h,zhu2026pursuing,zhu2026acl,zhu2026evaluating,zhao2026shaplora}, intent detection is usually solved as a classification task and slot filling as a sequence labeling task.


\begin{figure*}[t]
    \centering
    \includegraphics[width=\linewidth]{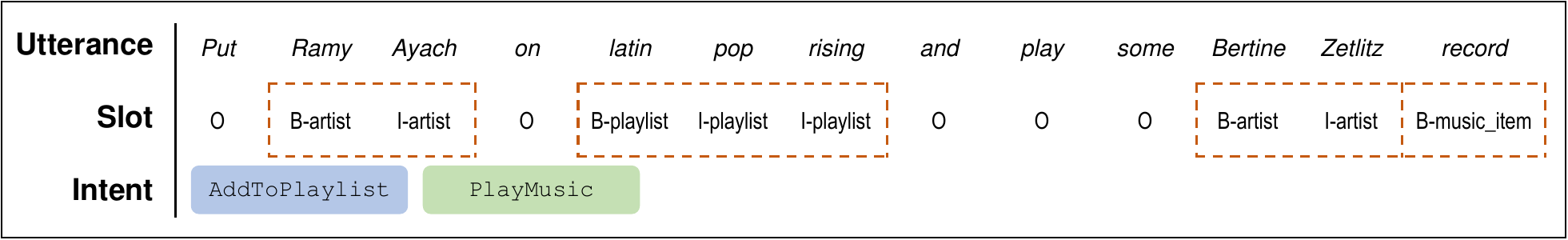}
    \caption{An example to illustrate the task, joint multiple intent detection and slot filling. We show the slot annotation in the \texttt{BIO} format, where \texttt{BIO} means Begin/Inside/Outside.}
    \label{fig:slu_example}
\end{figure*}

Since the two sub-tasks are strongly tied, current state-of-the-art systems typically adopt joint models to capture the interaction between the two sub-tasks\cite{DBLP:conf/ijcai/ZhangW16a,DBLP:conf/naacl/GooGHHCHC18,DBLP:conf/emnlp/LiLQ18,DBLP:conf/emnlp/QinCLWL19,Gunel2021SupervisedCL,tian2024opportunities,hersh2024search,tian2024opportunities,hersh2024search,zhu2024iapt,zhu-tan-2023-spt,Liu2022FewShotPF,xie2024pedro,Cui2023UltraFeedbackBL,zheng2024nat4at,zhu2023acf,gao2023f,zuo-etal-2022-continually,zhang-etal-2022-pcee,sun-etal-2022-simple,zhu-etal-2021-gaml,Zhu2021MVPBERTMP,li-etal-2019-pingan,zhu2019panlp,zhu2019dr,zhou2019analysis,zhang2025time,wang2025ts,liu2025parameter,yi2024drum,tian2024fanlora}. In particular, the intent prediction is usually leveraged to guide slot filling\cite{DBLP:conf/emnlp/QinXCL20}. Despite their success, most existing works only consider simple single intent scenarios, where each utterance only express a single intent. However, in the real world, users often express multiple intents in an utterance. For example, Gangadharaiah and Narayanaswamy\cite{DBLP:conf/naacl/GangadharaiahN19} show that 52\% of utterances in the amazon internal dataset have multiple intents. An example with multiple intent and slot annotation is shown in Figure~\ref{fig:slu_example}. The SLU system is required to detect all the intents (\emph{i.e.}, \texttt{AddToPlaylist} and \texttt{PlayMusic} in the example) and their corresponding slots.

Multi-intent SLU has posed challenges for both models and datasets. On the one hand, most existing single-intent SLU models cannot easily adapt to the multi-intent scenario. Though there are some neural-based models\cite{DBLP:conf/emnlp/QinXCL20,DBLP:conf/acl/QinWXXCL20,DBLP:journals/access/KumarB21} proposed in recent years for joint multiple intent detection and slot filling, the adopted interaction structure such as graph neural networks cannot take full advantage of pre-trained language models, and therefore limits the performance. On the other hand, there is currently a lack of high-quality multi-intent datasets. To this end, Qin \emph{et al.}\cite{DBLP:conf/emnlp/QinXCL20} constructed two multi-intent datasets, MixATIS and MixSNIPS, by randomly concatenating single-intent utterances in ATIS\cite{DBLP:conf/naacl/HemphillGD90} and SNIPS\cite{Coucke2018SNIPS}. Nevertheless, many samples in their constructed MixATIS and MixSNIPS are unnatural, which means the multiple intents expressed in the same utterance rarely co-occur in the real world.

This work is a response to the two challenges above. \textbf{For the challenge of models}, we propose a generative model, namely GEMIS, based on a pre-trained sequence-to-sequence language model, BART\cite{DBLP:conf/acl/LewisLGGMLSZ20}, to better capture the relation between the two sub-tasks with the pre-trained attention mechanism. In particular, we recast the joint multiple intent detection and slot filling as a sequence-to-sequence task, where the source sequence is the original utterance and the target sequence is a structured sequence of labels that contains intents, slot categories, and slot values. Besides, to better leverage the predictions of intents to guide the generation of slot categories and values, we design an attention-over-attention (AoA) structure to replace the cross-attention layers of BART decoder. \textbf{For the challenge of dataset}, we utilize pre-trained BERT\cite{DBLP:conf/naacl/DevlinCLT19} to construct two multi-intent datasets, namely MultiATIS and MultiSNIPS. Unlike previous work\cite{DBLP:conf/emnlp/QinXCL20} that randomly concatenates utterances with different intents to construct new ones, we utilize the next sentence prediction (NSP) head of BERT to obtain the probability that the concatenated utterances are successive sentences as the criterion of being selected into the constructed multi-intent datasets. Due to the usage of NSP, our constructed samples are more natural than those in previous work\cite{DBLP:conf/emnlp/QinXCL20}. Experimental results demonstrate that our proposed GEMIS achieves state-of-the-art performance on two public datasets, MixATIS and MixSNIPS, and our constructed datasets, MultiATIS and MultiSNIPS.

Our main contributions can be summarized as follows:
\begin{itemize}
    \item We propose a generative framework, \emph{i.e.}, GEMIS, for joint multiple intent detection and slot filling instead of solving them in classification and/or sequence labelling paradigm.
    \item We design an attention-over-attention (AoA) structure for BART decoder to guide slot filling with the attention pattern of intent predictions.
    \item With the next sentence prediction (NSP) head of BERT, we construct two high-quality multi-intent SLU datasets, \emph{i.e.}, MultiATIS and MultiSNIPS, which are comprised of 20,000 and 50,000 samples, respectively.
    \item Experimental results show that our proposed GEMIS achieves new state-of-the-art results on two public datasets, MixATIS and MixSNIPS, and also on our constructed datasets, MultiATIS and MultiSNIPS. Especially, our analysis demonstrates that the superiority of GEMIS is more significant as the number of intents in one utterance increases.
\end{itemize}

\section{Related Work}
\label{sec:related_work}
\subsection{Intent Detection}
Intent detection task can be solved as a multi-label text classification problem. Before 2012, the major area is to construct features to represent the semantics and syntax of sentence. For example, González-Caro and Baeza-Yates \cite{DBLP:conf/spire/Gonzalez-CaroB11} proposed multi-faceted query intent prediction to enhance the SVM classifier, Tür \emph{et al.}\cite{DBLP:conf/icassp/TurHHP11} proposed that adding simplified sentence structure to the features of AdaBoost model could improve the classification performance. After the rise of deep learning, researchers investigated neural-network-based classification models such as convolutional neural networks \cite{DBLP:conf/asru/XuS13} and recurrent neural networks \cite{DBLP:conf/interspeech/RavuriS15}, which refresh the state-of-the-art. Beyond the internal information (syntax and word context), external information was also applied in recent work. Firdaus \emph{et al.} \cite{DBLP:conf/pricai/FirdausBEB18} and Qiu \emph{et al.} \cite{DBLP:journals/access/QiuCJZ18} proposed to use pre-trained embedding such as Word2Vec\cite{DBLP:journals/corr/abs-1301-3781} and GloVe\cite{DBLP:conf/emnlp/PenningtonSM14} to improve the representation of utterance in GRU\cite{DBLP:journals/corr/ChungGCB14} and LSTM\cite{DBLP:conf/interspeech/SakSB14}. Capsule networks \cite{DBLP:conf/emnlp/XiaZYCY18} with zero-shot learning was proposed to discriminate the emerging intents via knowledge transfer.

\subsection{Slot Filling}
Slot filling is another critical task in natural language understanding, which attaches a slot label to each token in an utterance. As shown in  Figure~\ref{fig:slu_example}, we use the \texttt{BIO} tagging scheme to describe the slot labels in an utterance. Classical methods to solve sequence labelling task is using probabilistic graphical models including generative models and discriminative models. Generative models \cite{wang2005spoken} like hidden Markov model (HMMs) is to estimate the joint probability distribution of tokens and slot labels in an utterance, while discriminative models such as conditional random fields (CRFs)\cite{DBLP:conf/iccv/0001JRVSDHT15} is to learn the slot label condition probabilities when given the observed sequence. Similar to the 
intent detection task, deep learning also revolutionized this field. Yu \textit{et al.}\cite{DBLP:journals/jstsp/YuWD10} first combined the deep learning and CRF, proposing deep structured CRF. Mesnil \textit{et al.}\cite{DBLP:conf/interspeech/MesnilHDB13} and Kurata \textit{et al.} \cite{DBLP:conf/emnlp/KurataXZY16} proposed to use recurrent neural networks and LSTM for spoken language understanding. Gong \textit{et al.} \cite{DBLP:conf/aaai/GongLZOLZZDC19} proposed to jointly learn segment tagging, named entity tagging and slot filling, reaching state-of-the-art performance in slot filling at that time. Recently, Zhang \textit{et al.} \cite{DBLP:journals/symmetry/ZhangHW20} used deep time delay neural network from speech recognition that also performs state-of-the-art results in slot filling.

\subsection{Joint Intent Detection and Slot Filling}
Considering the close connection between intent detection and slot filling, the mass of research recently focus on the joint models. Qin \textit{et al.}\cite{DBLP:conf/ijcai/QinXC021} summarized existing methods of joint models as implicit joint modeling and explicit joint modeling. Implicit joint modeling shares the model encoder to capture the shared knowledge of utterance without explicit interaction between two sub-tasks. Zhang \textit{et al.}\cite{DBLP:conf/ijcai/ZhangW16a} proposed to use recurrent neural networks to learn the shared feature for intents and slots, Liu \textit{et al.}\cite{DBLP:conf/interspeech/LiuL16, DBLP:conf/sigdial/LiuL16} introduced bi-directional recurrent neural networks with attention and language modeling to improve the performance of joint intent detection and slot filling. Explicit joint modeling is to construct an interaction module between intents and slots, aiming to control the interaction process before performing prediction for specific tasks. Goo \textit{et al.}\cite{DBLP:conf/naacl/GooGHHCHC18} proposed the slot-gated mechanism in attention-based recurrent neural networks, realizing the conditional interaction between intents and slots. Qin \textit{et al.}\cite{DBLP:conf/emnlp/QinCLWL19} go a step further, using a stack-propagation framework to directly guide the slot filling with intent detection result, improving the state-of-the-art. E \textit{et al.}\cite{DBLP:conf/acl/ENCS19} proposed a bi-directional interrelated mechanism in the SF-ID network to enhance the two sub-tasks in one joint model. Qin \textit{et al.}\cite{DBLP:conf/icassp/QinLCKZ021} proposed a co-interactive transformer to strengthen the relationship between the two sub-tasks.

\section{Model}
\label{sec:method}

\par Our proposed \textbf{GE}nerative framework for \textbf{M}ulti-\textbf{I}ntent \textbf{S}LU (\textbf{GEMIS}) is introduced in this section. We give an overview of the framework in Section~\ref{subsec:sq2sq}, which is based on a pre-trained sequence-to-sequence language model introduced in Section~\ref{subsec:bart}. Further, we introduce the proposed attention-over-attention (AoA) structure in Section~\ref{subsec:aoa}, which is integrated in the decoder layers. At last, Section~\ref{subsec:training} describes the training process.

\subsection{Joint Multiple Intent Detection and Slot Filling as a Sequence-to-Sequence Task}
\label{subsec:sq2sq} 
\begin{figure}[t]
    \centering
    \includegraphics[width=\linewidth]{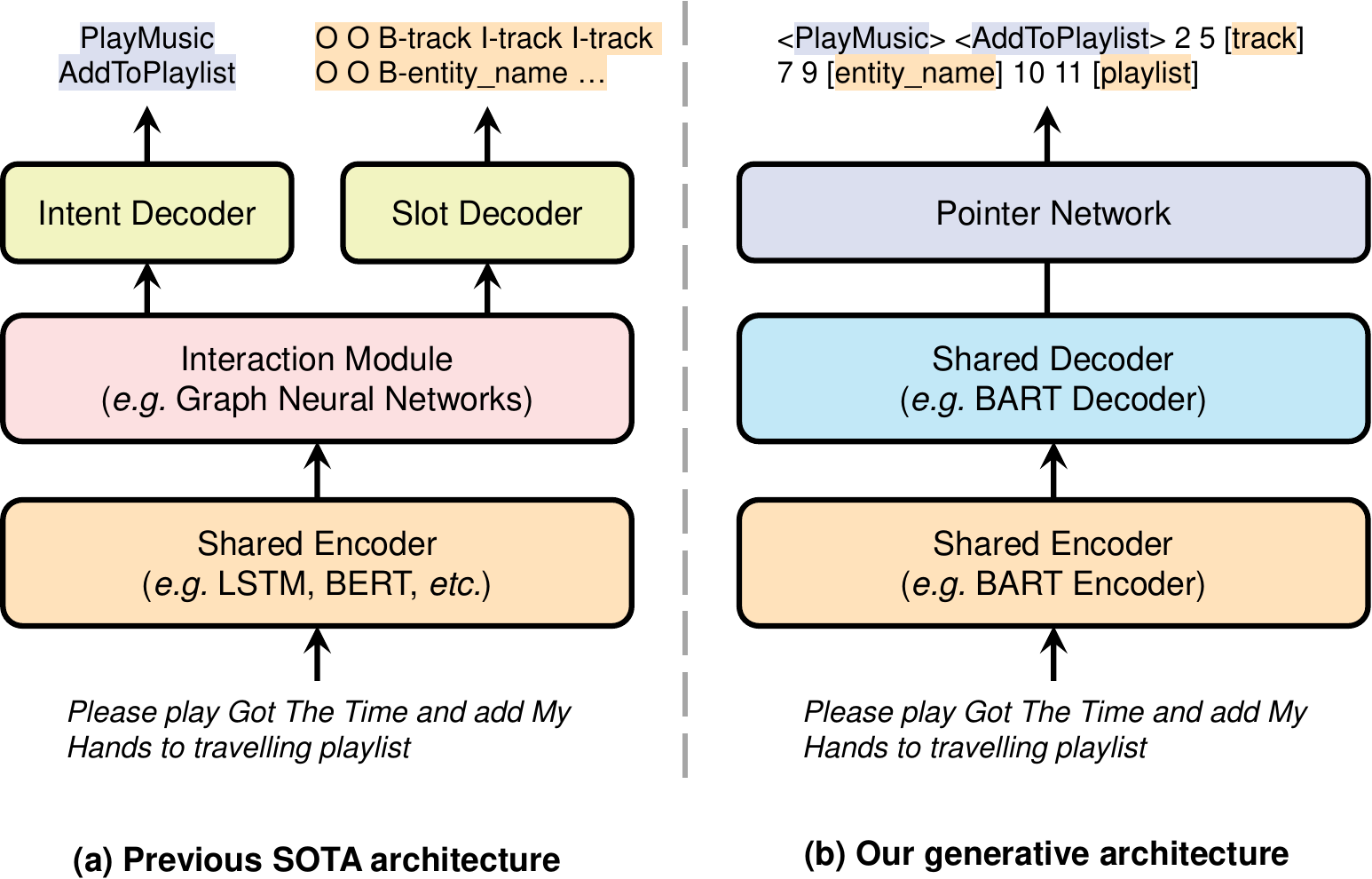}
    \caption{Comparison between previous state-of-the-art architectures (a) and our proposed generative architecture (b). In contrast to previous architectures where an interaction module is needed and trained from scratch, our architecture reformulates the two sub-tasks as a unified sequence-to-sequence task and uses a pre-trained shared decoder to capture the relationship between the two sub-tasks.}
    \label{fig:model}
\end{figure}

Sequence-to-sequence paradigm has shown great power and universality in various structured prediction tasks in language domain\cite{Sun2021Paradigm}, which motivates us to solve joint multiple intent detection and slot filling as a sequence-to-sequence task. In contrast to prior works that perform intent detection and slot filling with separate task-specific decoders, we adopt a shared decoder to generate intents and slots conditioned on the input utterance. Figure~\ref{fig:model} illustrates the difference between our method and previous state-of-the-art models. As shown in the figure, previous models usually employ an interaction module to capture the relationship between the two sub-tasks. Though the interaction modules are reasonable, they are trained from scratch and therefore can not fully enjoy the advantage of pre-trained parameters of language models such as BERT\cite{DBLP:conf/naacl/DevlinCLT19}. In our proposed generative framework, we adopt a pre-trained sequence-to-sequence language model, BART\cite{DBLP:conf/acl/LewisLGGMLSZ20}, to solve the joint task without additional randomly initialized parameters. As shown in Figure~\ref{fig:model}, we reformulate the intent and slot labels into a structured sequence. By this, the auto-regressive decoder can generate slot labels conditioning on the predicted intents, and therefore captures the interaction between intent detection and slot filling.

\begin{figure*}[t]
    \centering
    \includegraphics[width=\linewidth]{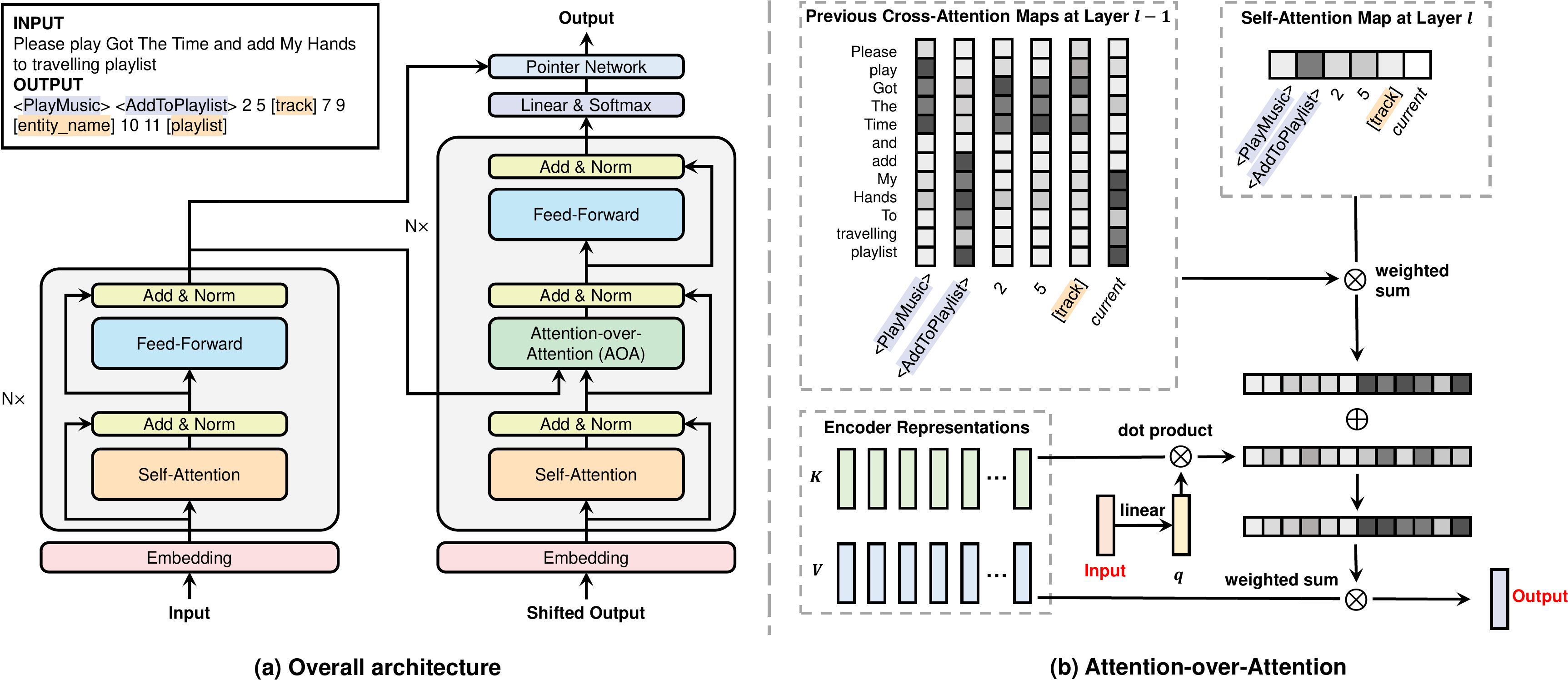}
    \caption{Illustration of the overall architecture and the attention-over-attention (AoA) structure.}
    \label{fig:aoa}
\end{figure*}

In particular, the intent and slot labels are reformulated as follows. Assume an utterance with $N$ tokens is denoted as $X = \langle x_1, x_2, ..., x_N\rangle$, the target sequence is its intent labels followed by its slot triplets, \emph{i.e.}, $\langle\text{Intent}_1, ..., \text{Intent}_p, s_1, e_1, \text{Slot}_1, ..., s_q, e_q, \text{Slot}_q\rangle$, where $s_i$ and $e_i$ are the start and end positions of the $i$-th slot, $p$ and $q$ are the number of intents and slots, respectively. Take the example from Figure~\ref{fig:model}, for the input utterance "\textit{Please play Got The Time and add My Hands to travelling playlist}", the target sequence is $\langle$\texttt{PlayMusic}, \texttt{AddToPlaylist}, 2, 5, \texttt{track}, 7, 9, \texttt{entity\_name}, 10, 11, \texttt{playlist}$\rangle$, where \texttt{PlayMusic} and \texttt{AddToPlaylist} are the intents expressed in the utterance, (2, 5, \texttt{track}), (7, 9, \texttt{entity\_name}), (10, 11, \texttt{playlist}) are the slot triplets. Take slot \texttt{track} for an instance, the filled value is indicated by the start position (\emph{i.e.}, 2) and the end position (\emph{i.e.}, 5), which represents "\textit{Got The Time}" in the utterance. Such a sequence-to-sequence paradigm can cope with variable number of intents and slots, and therefore is suitable for joint multiple intent detection and slot filling.

As a common practice for sequence-to-sequence tasks, we can use an auto-regressive decoder to generate the target sequence conditioning on the input sequence, \emph{i.e.},
\begin{equation}
    P(Y|X) = \prod_{t=1}^{M}P(y_t|X,Y_{<t}),
\end{equation}
where $X$ is the input sequence, $Y$ is the target sequence, $M$ is the number of tokens in the target sequence. Such a formulation can enjoy the power of pre-trained sequence-to-sequence language models such as BART\cite{DBLP:conf/acl/LewisLGGMLSZ20}, which will be introduced in the following subsection.

\subsection{The Modified BART Model}
\label{subsec:bart}
The BART model\cite{DBLP:conf/acl/LewisLGGMLSZ20} follows the standard Transformer architecture\cite{DBLP:conf/nips/VaswaniSPUJGKP17}, which consists a bidirectional encoder and a left-to-right decoder. BART is pre-trained with a denoising objective, which is to reconstruct the corrupted texts. We slightly modify BART to adapt to the joint task. Figure~\ref{fig:aoa}(a) gives an overall architecture of our modified BART, where we add a pointer network on the top and the cross-attention layers are replaced with our proposed attention-over-attention layers. As shown in the figure, the architecture contains an encoder and a decoder.

\textbf{Encoder} encodes the input utterance $X$ into vectors $\mathbf{H}^e$, \textit{i.e.},
\begin{equation}
\mathbf{H}^e = \text{Encoder}(X),
\end{equation}
where $\mathbf{H}^e\in \mathbb{R}^{N\times d}$, $N$ is the number of tokens in the input sequence, $d$ is the dimension of hidden vectors. The encoder is comprised of multiple Transformer encoder layers, each of which contains a self-attention sublayer and a feed-forward sublayer.

\textbf{Decoder} is to predict the intent or slot probability distribution $P(y_t|X, Y_{<t})$ for each step $t$. A special token \texttt{<SOS>} to indicate the start of the sequence is introduced as $y_0$. The generation process will stop when predicting another special token \texttt{<EOS>}, which means the end of the sequence. Considering that the tokens in the target sequence, which contains not only intent and slot categories but also position indices, are not in the BART vocabulary, it is necessary to make them readable to BART. For intent and slot categories such as \texttt{PlayMusic} and \texttt{track}, we directly add them to the BART vocabulary and initialize their embeddings as the mean vectors of their subwords. For position indicies, we employ a pointer network\cite{DBLP:conf/nips/VinyalsFJ15} to perform prediction over the input sequence. For the inputs of the decoder, we convert the previously predicted position indices to the corresponding wordpieces. Take the target sequence in Figure~\ref{fig:aoa} as an example, the predicted position 2 will be converted to its corresponding wordpiece "\textit{Got}" to be fed into the decoder. Denote the converted target sequence as $\Hat{Y}=\langle \hat{y}_1, \dots, \hat{y}_M\rangle$, the decoder can be formulated as follows,
\begin{align}
    \mathbf{h}^{d}_{t} &= \text{Decoder}(\mathbf{H}; \Hat{Y}_{<t}),\\
    \hat{y}_{t} &= \text{PointerNet}(\mathbf{h}^d_{t}, \mathbf{H}^e, \mathbf{E}_X),
\end{align}
where $\mathbf{h}^d_t$ is the decoding hidden state of the $t$-th token, $\mathbf{E}_X$ are the token embeddings of the input $X$, PointerNet is the pointer network. Similar to the encoder, the decoder is consisting of multiple decoder layers, each of which contains a self-attention sublayer, an attention-over-attention sublayer, and a feed-forward sublayer.

\textbf{Pointer network} is on the top of the decoder to perform position prediction over the input sequence, which is then fused with the prediction over intent and slot categories. The workflow of the pointer network is as follows,
\begin{align}
    \hat{\mathbf{H}}^e &= \alpha \cdot \text{MLP}(\mathbf{H}^e) + (1-\alpha)\cdot\mathbf{E}_X,\\
    \hat{\mathbf{y}}_t &= \text{Softmax}([\hat{\mathbf{H}}^e\otimes \mathbf{h}^d_t; \mathbf{W}\otimes \mathbf{h}^d_t]),\\
    \hat{y}_t &= \arg\max \hat{\mathbf{y}}_t,
\end{align}
where $\mathbf{E}_X\in\mathbb{R}^{N\times d}$ are the token embeddings of the input $X$, $\alpha$ is a hyper-parameter to balance the token embeddings and the encoder representations, $\otimes$ represents dot product, $[\cdot;\cdot]$ represents concatenating tensors, $\mathbf{W}\in\mathbb{R}^{L\times d}$ are the prediction weights over the intent and slot categories, $L$ is the total number of categories of intents and slots. As shown in the equations above, the encoder representations and the token embeddings are first fused into $\hat{\mathbf{H}}^e$, to be perform position predictions, which are then concatenated with intent and slot category predictions. The final prediction of the pointer network is performed over $N+L$ possible labels.

\subsection{Attention-over-Attention}
\label{subsec:aoa}
In original BART, the interaction between intent detection and slot filling is captured in the self-attention sublayers in the decoder. That is, the hidden states for generating current intents or slots can attend to previously generated intents and slots. Hence, such style of interaction lies in the branch of single flow interaction\cite{DBLP:conf/naacl/GooGHHCHC18,DBLP:conf/emnlp/QinCLWL19}, which allows slot filling to condition on the intent predictions. Further, to enhance the interaction between the two sub-tasks, we go beyond self-attention and design an attention-over-attention (AoA) sublayer to replace the original cross-attention sublayer in the BART decoder.

The general idea of AoA is to utilize the informative attention patterns of previously predicted intents and slots to guide the generation at current step. In particular, we manage to reuse the historical cross attention maps (CAMs) of previously generated labels. At generation step $t$, the CAM of the $l$-th layer is calculated as follows,
\begin{equation}
    \text{CAM}_t^l = \text{Softmax}(\mathbf{Q}_t^l\mathbf{K}_{enc}^{\top}),
\end{equation}
where $\mathbf{Q}_t^l\in\mathbb{R}^{t\times d}$ is the query from the decoder, $\mathbf{K}_{enc}\in\mathbb{R}^{N\times d}$ is the key from the encoder, and $N$ is the number of tokens in the input sequence. Intuitively, the CAMs can contain rich label-specific information that could help the prediction of related labels. Take the example from Figure~\ref{fig:aoa}, the CAM of the intent \texttt{PlayMusic} can highlight related tokens in the input sequence, \textit{e.g.}, "\emph{play Got The Time}", which would be helpful when generating position indices of the slot \texttt{track}.

In particular, at step $t$ and layer $l$, we weigh $\text{CAM}_t^l\in\mathbb{R}^{t\times N}$ with the self-attention maps (SAMs), which is 
\begin{equation}
    \text{SAM}_t^l = \text{Softmax}(\mathbf{q}_t^l\mathbf{K}_{dec}^{l\top}),
\end{equation}
where $\mathbf{q}_t^l\in\mathbb{R}^d$ is the query vector at step $t$ and layer $l$, $\mathbf{K}_{dec}^l\in\mathbb{R}^{t\times d}$ is the key matrix at the $l$-th decoder layer. Then we use the $\text{SAM}_t^l$ to weigh the $\text{CAM}_t^l$ to obtain a new cross-attention map that captures some prior knowledge encoded in the previous attention patterns. Since such an attention map is obtained by a self-attention map multiplied over a cross-attention map, we name this structure attention-over-attention (AoA). The obtained attention map is then combined with the original cross-attention map to calculate the final cross-attention map, \textit{i.e.,}
\begin{equation}
    \mathbf{A}_t^l = \text{Softmax}(\mathbf{q}_t^l\mathbf{K}_{enc}^{\top} + \frac{\text{SAM}_t^l\cdot\text{CAM}_t^l}{\sqrt{d_k}}),
\end{equation}
where $\mathbf{A}_t^l\in\mathbb{R}^{N}$, $\text{SAM}_t^l\in\mathbb{R}^t$, $\text{CAM}_t^l\in\mathbb{R}^{t\times N}$, and $d_k$ is the dimension of keys for each head. For simplicity, we neglect the calculation of multi-head attention in the equations above. The calculation of the AoA structure is depicted in Figure~\ref{fig:aoa}(b).

\subsection{Training and Inference}
\label{subsec:training}
By converting the joint multiple intent detection and slot filling task into a sequence-to-sequence task, we can use the conventional loss for training sequence-to-sequence models, \emph{i.e.},
\begin{equation}
    \mathcal{L} = -\frac{1}{M}\sum_{m=1}^{M}\sum_{i=1}^{N+L}\mathbf{y}_{m,i}\log{\hat{\mathbf{y}}_{m,i}},
\end{equation}
where $\mathbf{y}_m$ is the $m$-th one-hot ground truth in the target sequence and $\hat{\mathbf{y}}_m$ is the corresponding model prediction. During training, we use teacher forcing following common practice in training sequence-to-sequence models. During inference, we use greedy generation instead of beam search. Our pilot experiments show that beam search brings little effect on model performance, which is also found in \cite{DBLP:conf/acl/YanGDGZQ20}.

\section{Multi-Intent Dataset Construction}
\label{sec:newdata}

In real-world scenarios, users often express multiple intents in one utterance \cite{DBLP:conf/naacl/GangadharaiahN19}. However, most examples in widely used datasets such as ATIS\cite{DBLP:conf/naacl/HemphillGD90} and SNIPS\cite{Coucke2018SNIPS} only contain one intent. Qin \emph{et al.}\cite{DBLP:conf/emnlp/QinXCL20} constructed two synthetic multi-intent datasets, MixATIS and MixSNIPS, by randomly concatenating single-intent utterances in ATIS and SNIPS with some conjunctions such as "and then". Nevertheless, these randomly concatenated samples can be unnatural. For example, an utterance in MixATIS "\emph{Add Sara Hickman to my targeted list \textbf{and then} will the weather be stormy in aurora}" contains two intents that are almost impossible to co-occur in real-world scenarios. More examples can be found in Table~\ref{tab:cases}. In real-world task-oriented dialogue systems, users tend to experss multiple semantically related or progressive intents in the same utterance. To that end, we utilize the next sentence prediction (NSP) head of BERT\cite{DBLP:conf/naacl/DevlinCLT19} to construct more realistic multi-intent datasets, namely MultiATIS and MultiSNIPS.

\begin{table*}[t]
    \centering
    \caption{Examples in MixATIS/MixSNIPS and MultiATIS/MultiSNIPS.}
    \label{tab:cases}
    \resizebox{\linewidth}{!}{
    \begin{tabular}{l}
    \toprule
    \textbf{Examples in MixATIS and MixSNIPS}    \\ \midrule
    I need a ticket from Nashville to Seattle \emph{and then} flight numbers from Chicago to Seattle on continental.   \\
    Tell me about the m80 aircraft, list airports \emph{and then} how many Canadian airlines flights use aircraft dh8. \\
    List California airports, which flights travel from Cleveland to Indianapolis on April fifth \emph{and also} what are the fairs for ground transportation in Denver?   \\
    Add Sara Hickman to my targeted list \emph{and then} will the weather be stormy in aurora.	 \\ 
    This track should go into my playlist called this is Beethoven , I want to eat mezes at the pub for 1 at four pm \emph{and} find me haunted castle.	\\
    \midrule
    \textbf{Examples in MultiATIS and MultiSNIPS}    \\ \midrule
    I would like to fly from Boston to Baltimore \emph{and} show me ground transportation in Boston and in Baltimore.  \\
    Which flights on us air go from Orlando to Cleveland \emph{and} I'm trying to find the flight number from a flight from Orlando to Cleveland on us air and it arrives around 10 pm.  \\
    List the number of flights leaving Boston for Dallas fort worth before 9 am in the morning \emph{and then} what is the earliest flight from Boston to Dallas fort worth leaving august eighth? \\
    I want to eat in 19 hours at NM \emph{and} what will the weather be in NM in 1 minute.	\\
    I'd like for you to put this artist to my evening commute playlist \emph{and then} play my playlist tgif on iTunes.	\\
    \bottomrule
    \end{tabular}
    }
\end{table*}

\renewcommand{\algorithmicrequire}{\textbf{Input:}}
\renewcommand{\algorithmicensure}{\textbf{Output:}}
\removelatexerror
 \begin{algorithm}
	\caption{Multi-Intent Dataset Construction}
	\label{alg:dataset}
	\begin{algorithmic}[1]
		\renewcommand{\algorithmicrequire}{\textbf{Input:}}
		\renewcommand{\algorithmicensure}{\textbf{Output:}}
		\REQUIRE Single-intent dataset $U_s$, coherence threshold $\tau$
		\ENSURE  Multi-intent dataset $U_m$
		\STATE Initialize multi-intent dataset $U_m\gets \varnothing$
		\FOR {$u_s$ in $U_s$}
		    \STATE Sample the number of intents $n\in\{1, 2, 3\}$ with probabilities $(0.3, 0.5, 0.2)$
		    \STATE Initialize the multi-intent utterance $u_m \gets u_s$
		    \WHILE{ $n$ > 1}
		    \STATE Filter out utterances in $U_s$ with intents appeared in $u_m$ and obtain a candidate set $\hat{U}_s$
		    \FOR {$u_c$ in $\hat{U}_s$}
		        \IF{ $P_{\text{NSP}}(u_m||u_c) > \tau$ }
		            \STATE $u_m\gets u_m||u_c$ \ \ \ \ $\triangleright$ $||$ \textit{means concatenation}
		            \STATE \textbf{break}
		        \ENDIF
		    \ENDFOR
	        \STATE $n \gets n - 1$
		    \ENDWHILE
		    \STATE Put $u_m$ into $U_m$
		\ENDFOR
		\RETURN Multi-intent dataset $U_m$
	\end{algorithmic}
\end{algorithm}

\begin{figure*}[htbp]
\centering
\includegraphics[width=.49\linewidth]{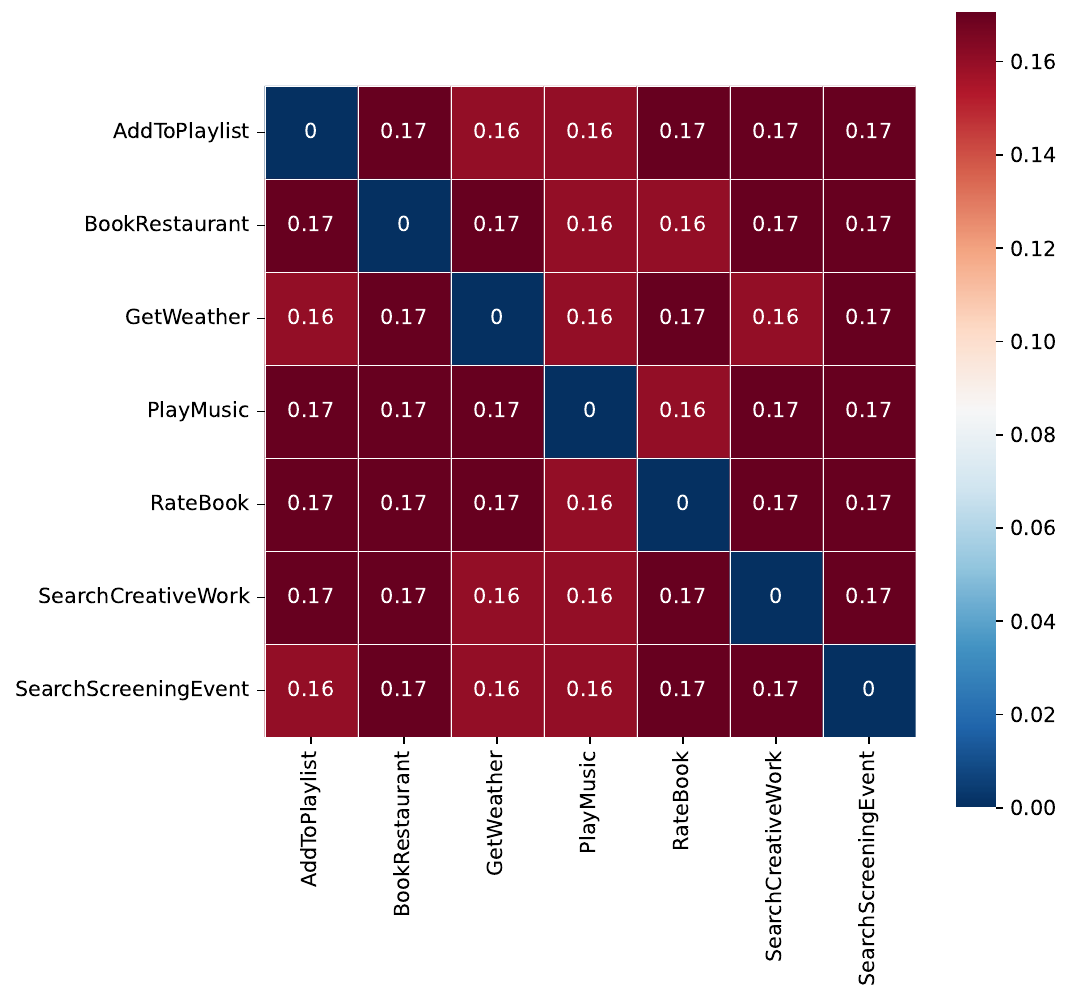}
\includegraphics[width=.49\linewidth]{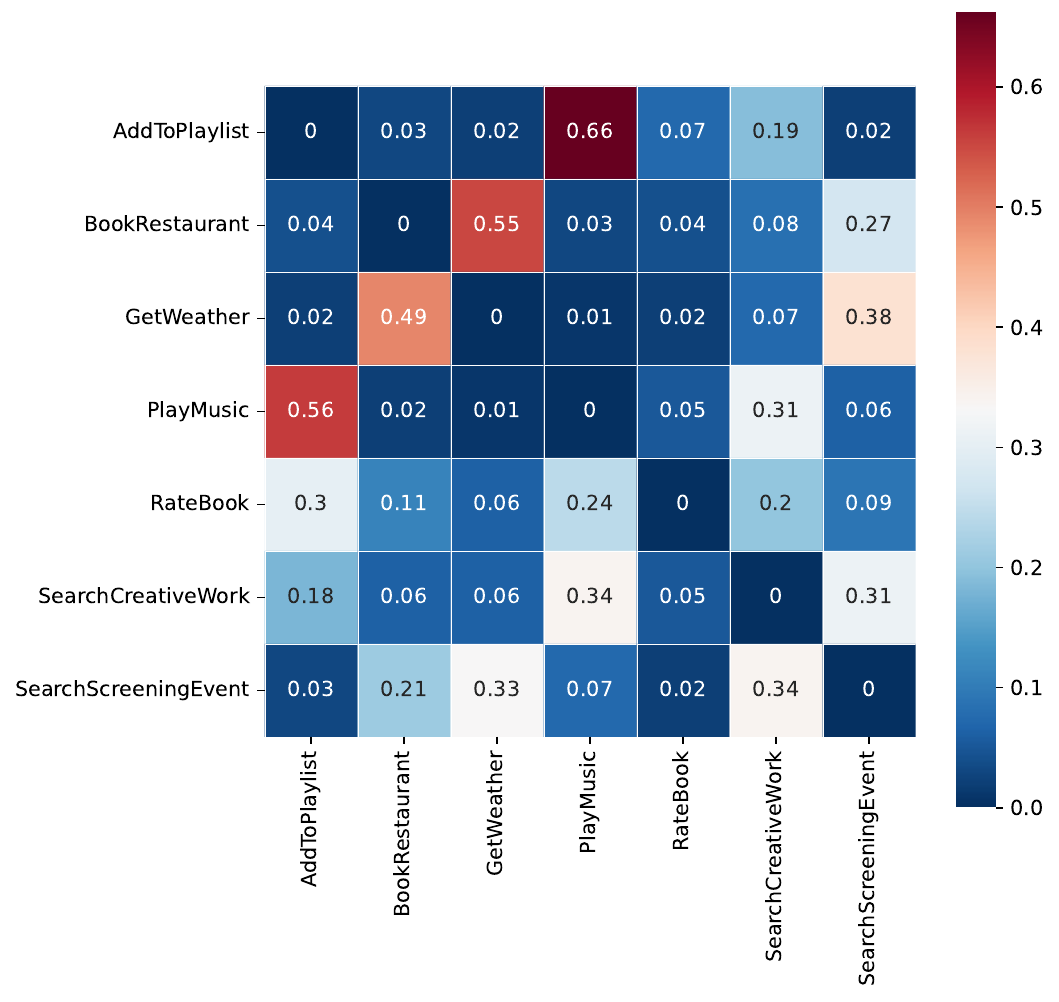}
\caption{Intent co-occurrence of MixSNIPS and MultiSNIPS. (a) In MixSNIPS~\cite{DBLP:conf/emnlp/QinXCL20}, the distribution of intent co-occurrence for each intent is a uniform distribution because MixSNIPS randomly concatenates single-intent utterances regardless of the relationship between the intents. (b) In our constructed MultiSNIPS, related intents, \textit{e.g.}, \texttt{AddToPlaylist} and \texttt{PlayMusic}, \texttt{BookRestaurant} and \texttt{GetWeather}, frequently appear in the same utterance, which is more realistic.}
\label{fig:co-occurence}
\end{figure*}

\subsection{Constructing Multi-Intent Datasets with Next Sentence Prediction}
Next Sentence Prediction (NSP) is a pre-training task used in BERT\cite{DBLP:conf/naacl/DevlinCLT19}, which is to predict whether a pair of sentences are successive sentences. In particular, BERT chooses two sentences \texttt{A} and \texttt{B} for each pre-training example, meanwhile 50\% of the time \texttt{B} is the actual next sentence of \texttt{A} and 50\% of the time \texttt{B} is a random sentence sampled from the corpus. During pre-training, BERT is trained to predict whether \texttt{B} is the next sentence of \texttt{A}. After pre-training, the NSP head of BERT can effectively capture the relationship of a pair of sentences, and therefore is suitable for scoring the coherence of two utterances, which can be regarded as a criterion of concatenating. Hence, we define the coherence score $P_{coh}$ of two utterances based on the next sentence probability,
\begin{equation}
    P_{coh} = P_{\text{NSP}}(u_A||u_B),
\end{equation}
where $u_A$ and $u_B$ are two utterances, $||$ means string concatenation. If the coherence score exceeds a pre-defined threshold, we can concatenate the two utterances as a single utterance (\emph{e.g.}, "$u_A$ \textit{and then} $u_B$") that expresses multiple related intents. If the coherence score does not exceed the threshold, we filter out the concatenated sample. By iteratively performing such concatenate-evaluate process, we can also generate samples that have more than two intents. Algorithm \ref{alg:dataset} shows the detalied process of constructing multi-intent datasets. Similar with Qin \emph{et al.}~\cite{DBLP:conf/emnlp/QinXCL20}, we construct our synthetic datasets MultiATIS and MultiSNIPS based on ATIS and SNIPS.

\subsection{Comparison with MixATIS and MixSNIPS}
Rather than randomly concatenating single-intent utterances like MixATIS and MixSNIPS, we construct our MultiATIS and MultiSNIPS by concatenating single-intent utterances that have high coherence scores, which is expected to result in higher quality of datasets. However, comprehensively evaluating the quality of datasets can be labor-intensive. 

In this section, we verify the quality of our constructed MultiATIS and MultiSNIPS by counting the frequency of intent co-occurrence. Ideally, users tend to express related or progressive intents in the same utterance, so a high-quality multi-intent dataset is expected to show high co-occurrence of related intents. Figure~\ref{fig:co-occurence} shows the distributions of intent co-occurrence for MixSNIPS~\cite{DBLP:conf/emnlp/QinXCL20} and our constructed MultiSNIPS. As shown in Figure~\ref{fig:co-occurence}(a), the distribution of intent co-occurrence for each intent in MixSNIPS is a uniform distribution since it does not consider the relationship among these intents when concatenating utterances. In our constructed MultiSNIPS, as shown in Figure~\ref{fig:co-occurence}(b), the distribution of intent co-occurrence for each intent is quite different. For example, the intent \texttt{AddToPlaylist} tends to appear with \texttt{PlayMusic} in the same utterance; the intent \texttt{BoookRestaurant} tends to appear with \texttt{GetWeather} in the same utterance. Thus, the intent co-occurrence of MultiSNIPS is more intuitive and realistic than that of MixSNIPS. Some examples of our constructed datasets can be found in Table~\ref{tab:cases}.

\section{Experiments}
\label{sec:exp}
\subsection{Datasets and metrics}
We first conduct experiments on two public multi-intent datasets, MixATIS and MixSNIPS, which is constructed by Qin \textit{et al.}\cite{DBLP:conf/emnlp/QinXCL20}. MixATIS is created by randomly concatenating samples in the ATIS dataset\cite{DBLP:conf/naacl/HemphillGD90}. MixSNIPS is constructed based on the SNIPS dataset\cite{Coucke2018SNIPS} in the same way of MixATIS. For MixATIS and MixSNIPS, the number of intents in one utterance ranges from 1 to 3, with proportions of $[0.3, 0.5, 0.2]$, respectively. We use the clean versions of MixATIS and MixSNIPS that remove the repeated utterances in original datasets\footnote{\url{https://github.com/LooperXX/AGIF}}. Besides, we also conduct experiments on our constructed MultiATIS and MultiSNIPS (Section \ref{sec:newdata}), which are also based on ATIS and SNIPS but consider the relationship between the utterances to be concatenated. The statistics of our evaluated datasets are shown in Table \ref{table:dataset}.

\begin{table}[ht]
\centering
\caption{Statistics of our used datasets.}
\label{table:dataset}
\resizebox{0.49\textwidth}{!}{
\begin{tabular}{ccccc}
\toprule
\textbf{Dataset} & \textbf{MixATIS} & \textbf{MultiATIS} & \textbf{MixSNIPS} & \textbf{MultiSNIPS} \\ \midrule
Train   & 13162   & 18000     & 39776    & 45000      \\
Dev     & 759     & 1000      & 2198     & 2500       \\
Test    & 828     & 1000      & 2199     & 2500       \\ \bottomrule
\end{tabular}}
\end{table}

To evaluate the system performance on joint multiple intent detection and slot filling, we follow the metrics used in prior work\cite{DBLP:conf/naacl/GooGHHCHC18,DBLP:conf/emnlp/QinCLWL19}. We evaluate the performance of slot filling using F1 score, intent detection using accuracy, the sentence-level semantic frame parsing using overall accuracy, which represents all of the intents and slots in the utterance are correctly predicted. We use the overall accuracy as the primary metric to perform model selection.

\subsection{Implementation details}
The parameters and configuration of our sequence-to-sequence model are initialized with BART\textsubscript{LARGE}\cite{DBLP:conf/acl/LewisLGGMLSZ20}. Our implementation is based on Hugginface's Transformers~\cite{Wolf2019Huggingface}. We use AdamW\cite{DBLP:conf/iclr/LoshchilovH19} with batch size of 16 to optimize our model for 30 epochs on each dataset. We perform hyper-parameter search with learning rate in \{1e-5, 2e-5, 3e-5\} and select the models that achieve the best performance on the development sets. All the experiments are conducted with a single NVIDIA Tesla V100 SXM2.

\begin{table*}[ht]
\centering
\caption{Main results on MixATIS and MixSNIPS.}
\label{table:main}
\newcommand{\tabincell}[2]{\begin{tabular}{@{}#1@{}}#2\end{tabular}}
\resizebox{0.99\textwidth}{!}{
\begin{tabular}{l||ccc||ccc}
\hline
\multirow{2}{*}{\textbf{Models}} & \multicolumn{3}{c||}{\textbf{MixATIS}}          & \multicolumn{3}{c}{\textbf{MixSNIPS}}          \\ \cline{2-7} 
                        & Slot (F1) & Intent (Acc) & Overall (Acc) & Slot (F1) & Intent (Acc) & Overall (Acc) \\ \hline
Slot-Gated\cite{DBLP:conf/naacl/GooGHHCHC18}     & 87.7     & 63.9        & 35.5         & 87.9     & 94.6        & 55.4         \\
SF-ID\cite{DBLP:conf/acl/ENCS19}   & 87.4     & 66.2        & 34.9         & 90.6     & 95.0        & 59.9         \\
Stack-Propagation      & 87.8     & 72.1        & 40.1         & 94.2     & 96.0        & 72.9         \\
Joint Multiple ID-SF    & 84.6     & 73.4        & 36.1         & 90.6     & 95.1        & 62.9         \\
AGIF\cite{DBLP:conf/emnlp/QinXCL20}    & 86.7     & 74.4        & 40.8         & 94.2     & 95.1        & 74.2         \\
GL-GIN\cite{DBLP:conf/acl/QinWXXCL20}               & 88.3     & 76.3        & 43.5         & 94.9     & 95.6        & 75.4         \\ \hdashline
GEMIS w/o AoA                    & 88.6     & 80.0        & 52.7         & 96.7     & 97.3        & 85.1         \\
GEMIS                     & \textbf{89.2}     & \textbf{81.4}        & \textbf{53.4}         & \textbf{97.4}     & \textbf{98.1}        & \textbf{87.4}         \\ \hline
\end{tabular}}
\end{table*}

\subsection{Baselines}
We compare our proposed model with existing state-of-the-art multi-intent SLU baselines:
\begin{itemize}
  \item \textbf{Slot-Gated}\cite{DBLP:conf/naacl/GooGHHCHC18}, a slot-gated model that explicitly utilizes the correlation of intent detection and slot filling.
  \item \textbf{SF-ID}\cite{DBLP:conf/acl/ENCS19}, which directly connects intent detection and slot filling, promoting each other mutually.
  \item \textbf{Stack-Propagation}\cite{DBLP:conf/emnlp/QinCLWL19}, a joint model that performs token-level intent detection and captures intent semantic knowledge and alleviates error propagation from intent.
  \item \textbf{Joint Multiple ID-SF}\cite{DBLP:conf/naacl/GangadharaiahN19}, an attention-based neural network that performs joint multiple intent detection and slot filling at token-level.
  \item \textbf{AGIF}\cite{DBLP:conf/emnlp/QinXCL20}, an adaptive graph-interaction framework to realize the fine-grained integration between multiple intent detection and slot filling.
  \item \textbf{GL-GIN}\cite{DBLP:conf/acl/QinWXXCL20}, a non-autoregressive model with local slot-aware and global intent-slot graph interaction layer to alleviate the uncoordinated slots problem and enhance the interaction of intent detection and slot filling.
\end{itemize}

\begin{table*}[ht]
\centering
\caption{Main results on our constructed MultiATIS and MultiSNIPS.}
\label{table:new}
\resizebox{0.99\textwidth}{!}{
\newcommand{\tabincell}[2]{\begin{tabular}{@{}#1@{}}#2\end{tabular}}
\begin{tabular}{l||ccc||ccc}
\hline
\multirow{2}{*}{\textbf{Models}} & \multicolumn{3}{c||}{\textbf{MultiATIS}}        & \multicolumn{3}{c}{\textbf{MultiSNIPS}}        \\ \cline{2-7} 
                        & Slot (F1) & Intent (Acc) & Overall (Acc) & Slot (F1) & Intent (Acc) & Overall (Acc) \\ \hline
AGIF\cite{DBLP:conf/emnlp/QinXCL20}                   & 94.3     & 68.2        & 60.2         & 94.3     & 95.9        & 73.9         \\
GL-GIN\cite{DBLP:conf/acl/QinWXXCL20}                 & 94.8     & 74.5        & 58.3         & 93.6     & 64.9        & 53.0         \\ \hdashline
GEMIS w/o AoA                    & 95.2     & 93.4        & 71.1         & 97.6     & 98.5        & 88.9         \\
GEMIS                     & \textbf{95.4}     & \textbf{94.1}        & \textbf{71.5}         & \textbf{98.1}     & \textbf{98.8}        & \textbf{91.5}         \\ \hline
\end{tabular}}
\end{table*}

\subsection{Main results}
Experimental results on MixATIS and MixSNIPS are shown in Table \ref{table:main}. Our proposed GEMIS achieves new state-of-the-art performances in terms of the considered metrics on the two datasets. On both MixATIS and MixSNIPS, GL-GIN achieves the best overall accuracy (43.5\% and 75.4\%) among the baseline models. In contrast, our proposed GEMIS achieves 53.4\% and 87.4\% overall accuracy, outperforming all of the baseline models by a large margin. Without AoA, GEMIS can still achieve 52.7\% and 85.1\% overall accuracy, demonstrating the superior performance of the generative framework. By replacing conventional cross-attention with our proposed AoA structure, we obtain improvement on all the metrics (slot F1, intent accuracy, and overall accuracy) on the both datasets. Especially, AoA achieves 2.3\% absolute points improvement of overall accuracy on MixSNIPS over vanilla GEMIS.

Experimental results on our constructed MultiATIS and MultiSNIPS are shown in Table~\ref{table:new}. We compare our proposed GEMIS with two state-of-the-art models, namely AGIF\cite{DBLP:conf/emnlp/QinXCL20} and GL-GIN\cite{DBLP:conf/acl/QinWXXCL20}. Similar with experimental results on MixATIS and MixSNIPS, GEMIS significantly outperforms previous methods on MultiATIS and MultiSNIPS. GEMIS achieves 11.3\% and 17.6\% absolute point improvement on overall accuracy of MultiATIS and MultiSNIPS, respectively. The improvements over previous state-of-the-art models on MultiATIS and MultiSNIPS are more significant than that on MixATIS and MixSNIPS, indicating that GEMIS is more favorable in real-world dialogue scenarios.

\subsection{On number of intents}
Each sample in our evaluated multi-intent datasets contains variable number of intents (range from 1 to 3). In this section we evaluate model performance on sub-datasets with different number of intents. In particular, we divide each test set into three sub-sets by the number of intents and evaluate GL-GIN, AGIF, and our proposed GEMIS on each sub-set. Figure \ref{fig:acc} shows the trend of overall accuracy as the number of intents increases. We observe that:

\begin{figure*}[htbp]
\centering
\includegraphics[width=.4\linewidth]{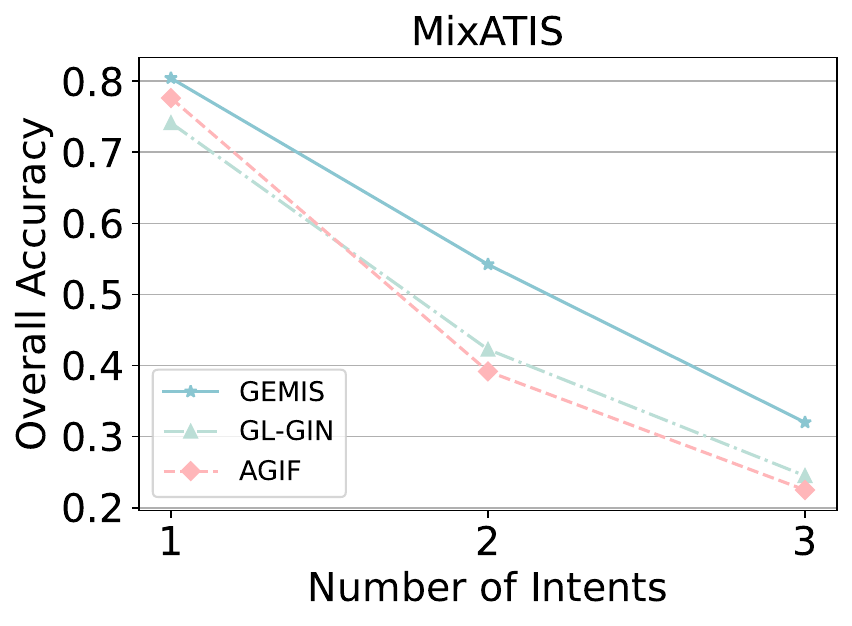}
\includegraphics[width=.4\linewidth]{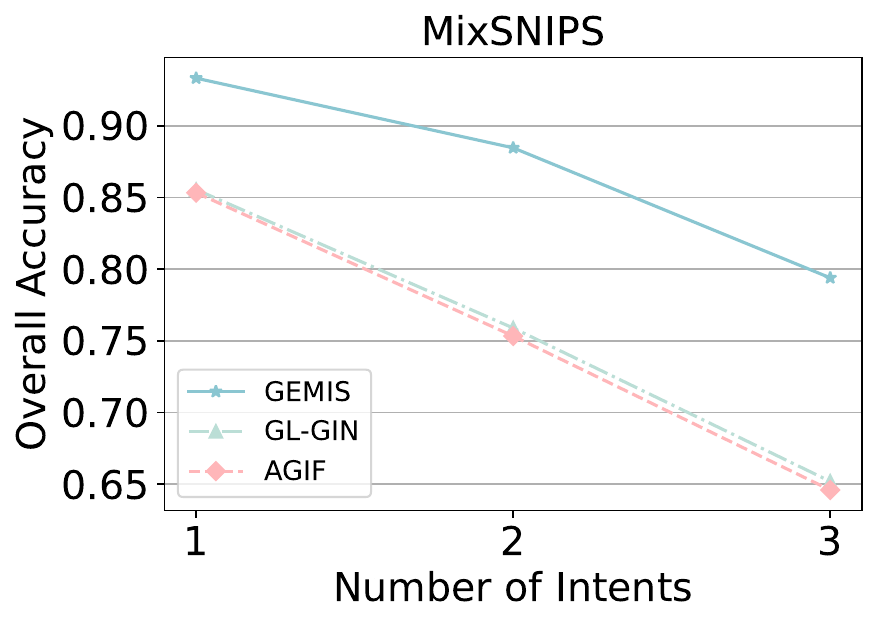}
\\
\includegraphics[width=.4\linewidth]{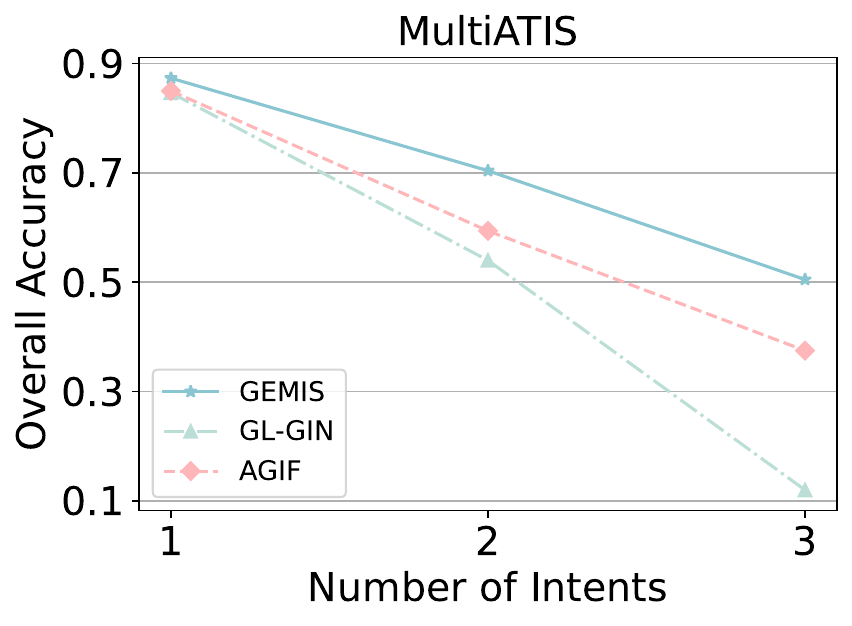}
\includegraphics[width=.4\linewidth]{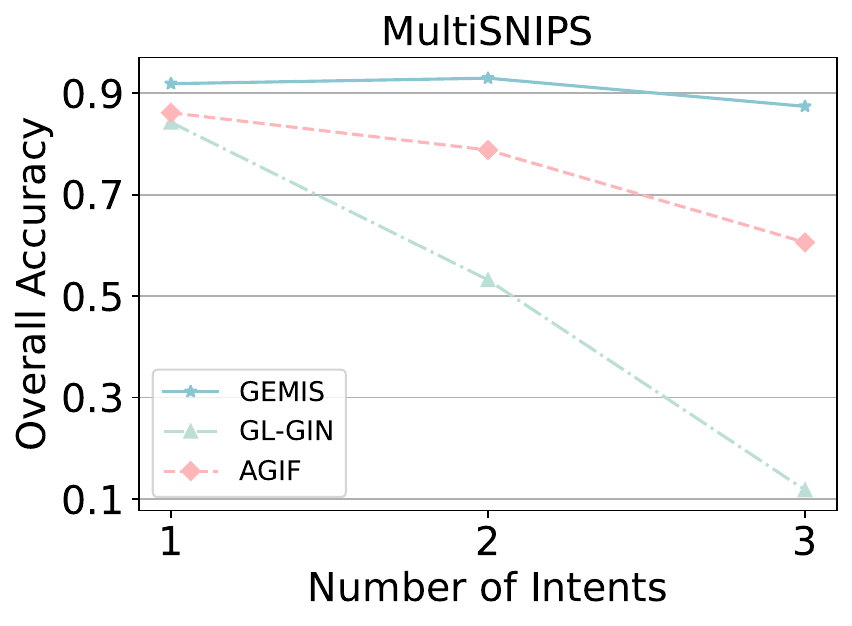}
\caption{Overall accuracy with different number of intents. Our proposed GEMIS significantly outperforms previous state-of-the-art methods as the number of intents increases.
}
\label{fig:acc}
\end{figure*}

\begin{table*}[ht]
\centering
\caption{Results of GEMIS using BART\textsubscript{BASE} and BART\textsubscript{LARGE}.}
\label{table:bart_base}
\newcommand{\tabincell}[2]{\begin{tabular}{@{}#1@{}}#2\end{tabular}}
\begin{tabular}{c||ccc||ccc}
\hline
\multirow{2}{*}{\textbf{Dataset}} & \multicolumn{3}{c||}{GEMIS\textsubscript{BASE}}      & \multicolumn{3}{c}{GEMIS\textsubscript{LARGE}}      \\ \cline{2-7} 
                         & Slot (F1) & Intent (Acc) & Overall (Acc) & Slot (F1) & Intent (Acc) & Overall (Acc) \\ \hline
MixATIS                  & 88.5     & 81.3        & 52.5         & 89.2     & 81.4        & 53.4         \\
MixSNIPS                 & 96.6     & 96.5        & 84.3         & 97.4     & 98.1        & 87.4         \\
MultiATIS                & 94.9     & 94.7        & 69.5         & 95.4     & 94.1        & 71.5         \\
MultiSNIPS               & 97.3     & 98.3        & 88.2         & 98.1     & 98.8        & 91.5         \\ \hline
\end{tabular}
\end{table*}

\begin{itemize}
    \item{On single-intent samples, both GEMIS and the baseline models achieve competitive performance, in which case GEMIS only slightly outperforms its counterparts.}
    \item{As the number of intents increases, the overall accuracy of both GEMIS and baseline models decreases, which demonstrates the challenge of multiple intents. Nevertheless, in contrast to baseline methods, GEMIS shows less performance degradation on multi-intent samples. In other words, the performance superiority of GEMIS over baseline models is enlarged as the number of intents increases.}
    \item{The performance degradation of GEMIS on MultiATIS and MultiSNIPS is not as significant as it is on MixATIS and MixSNIPS, indicating that the relationship of the different intents can be well exploited by GEMIS to help prediction. However, in MixATIS and MixSNIPS, one utterance often carries unrelated intents, resulting in severe performance degradation.}
\end{itemize}

\subsection{Effect of model size}
Typically, the size of the pre-trained language model has an important effect on the performance. In previous experiments, our evaluated GEMIS is built on BART\textsubscript{LARGE} by default. In this section, to study the effect of model size on the four datasets, we also conduct experiments using BART\textsubscript{BASE}. We denote GEMIS using BART\textsubscript{LARGE} (BART\textsubscript{BASE}) as GEMIS\textsubscript{LARGE} (GEMIS\textsubscript{BASE}). As shown in Table \ref{table:bart_base}, GEMIS\textsubscript{LARGE} outperforms GEMIS\textsubscript{BASE}, which confirms the increasing power of large-scale pre-trained language models. Nevertheless, GEMIS\textsubscript{BASE} still outperforms previous baseline models by a large margin, indicating the effectiveness of our generative framework for joint multiple intent detection and slot filling.

\section{Conclusion}
\label{sec:conclusion}
In this work, we formulate the joint multiple intent detection and slot filling as a unified sequence-to-sequence task, and use a pre-trained sequence-to-sequence model, BART~\cite{DBLP:conf/acl/LewisLGGMLSZ20}, to handle it. In particular, we build a pointer network on the top of BART to perform joint prediction over positions of input sequences and categories of intents and slots. Besides, we propose a novel attention-over-attention (AoA) module to replace the cross-attention in the BART decoder, which incorporates an inductive bias to capture the interaction between the intent detection and slot filling. In addition, we propose to leverage the next sentence prediction (NSP) head of BERT~\cite{DBLP:conf/naacl/DevlinCLT19} to construct two new multi-intent datasets based on single-intent datasets. Experimental results demonstrate that our proposed model achieves state-of-the-art performance on two public multi-intent datasets and our constructed multi-intent datasets.

\bibliography{custom}
\bibliographystyle{acl_natbib}

\appendix

\end{document}